\documentclass[aps,reprint,floatfix,superscriptaddress,amsmath,amssymb]{revtex4-2}

\usepackage{graphicx}
\usepackage{booktabs}
\usepackage{amsfonts}
\usepackage{mathtools}
\usepackage{amsthm}
\usepackage{bm}
\usepackage{array}
\usepackage{xcolor}
\usepackage{listings}
\usepackage{url}
\newcommand{\hquad}{\hspace{0.5em}}

\definecolor{predblue}{RGB}{35,112,187}
\definecolor{labelgray}{RGB}{145,145,145}
\definecolor{latentred}{RGB}{242,101,91}
\definecolor{tokenblue}{RGB}{190,226,241}

\usepackage[most]{tcolorbox}
\tcbuselibrary{listings}
\newtcblisting{promptbox}[1][]{
  listing only,
  enhanced,
  width=\textwidth,
  colback=gray!6,
  colframe=gray!30,
  coltitle=black,
  arc=1.5mm,
  boxrule=0.4pt,
  left=1.2mm,
  right=1.2mm,
  top=0.8mm,
  bottom=0.8mm,
  boxsep=1pt,
  toptitle=1.0mm,
  bottomtitle=0.4mm,
  fonttitle=\bfseries\footnotesize,
  title={\strut #1},
  titlerule=1pt,
  listing options={
    basicstyle=\ttfamily\scriptsize,
    breaklines=true,
    breakatwhitespace=false,
    columns=fullflexible,
    keepspaces=false,
    showstringspaces=false,
    breakindent=0pt,
    breakautoindent=false,
    xleftmargin=0pt,
    framexleftmargin=0pt
  }
}

\begin{document}

\title{Uncovering Spontaneous Physics Representations in In-Context Learning}

\author{Yeongwoo Song}
\thanks{These authors contributed equally to this work.}
\author{Jaeyong Bae}
\thanks{These authors contributed equally to this work.}
\affiliation{Department of Physics, KAIST, Republic of Korea}

\author{Dong-Kyum Kim}
\affiliation{MPI for Security and Privacy, Germany}

\author{Hawoong Jeong}
\email{hjeong@kaist.edu}
\affiliation{Department of Physics, KAIST, Republic of Korea}
\affiliation{Center for Complex Systems, KAIST, Republic of Korea}

% \date{\today}

\begin{abstract}
In-context learning (ICL) lets large language models (LLMs) solve new tasks from prompts alone, across an ever-widening range of domains, yet the mechanisms underlying this ability remain poorly understood.
Physical systems offer a controlled testbed for this question as they provide experimentally controllable data with structured dynamics grounded in fundamental principles.
Here we study the ICL ability of LLMs, focusing on physical reasoning.
Using dynamics forecasting as a proxy task, we first show that LLMs forecast physical dynamics in context, with accuracy improving as more history is provided.
Analyzing the model's residual stream reveals internal activations that correlate with key physical quantities such as energy.
These correlations strengthen gradually with context length, indicating that LLMs spontaneously form representations aligned with physical concepts without any physics-specific supervision.
To assess whether these representations contribute to the model's predictions, we introduce a layer-wise gradient-based attribution analysis.
We find that, residual directions more strongly correlated with energy also receive greater attribution to numerical predictions.
This pattern is not observed for features correlated with directly observed quantities such as displacement, suggesting that the energy-related signal is not merely numerical information copied from the input.
Our results broaden ICL analysis to structured physical dynamics and give a mechanistic account of how LLMs organize physical structure in context.
\end{abstract}

\maketitle

\section{Introduction}
\label{sec.1}
From the advent of transformers~\cite{vaswani2017attention} to the development of large language models (LLMs), recent advances in artificial intelligence (AI) have produced highly capable systems that are now widely deployed across daily life and society. Among the diverse capabilities of these systems, an especially striking one is the emergent ability to perform complex tasks in context, known as in-context learning (ICL). This ability goes far beyond their original training objective of next-token prediction on text~\cite{brown2020language, wei2022emergent, bubeck2023sparks}. ICL has since been demonstrated across diverse classes of tasks, including basic arithmetic, reasoning, function learning, reinforcement learning, and time-series prediction~\cite{webb2023emergent, garg2022can, brooks2022context, gruver2023large}.
A particularly notable instance is the ability to extrapolate the behavior of dynamical systems~\cite{liu2024llms}.
Such capability suggests that LLMs might internally represent and manipulate physics knowledge such as laws of motion or conservation principles.

Nevertheless, the specific internal representations and mechanisms that enable LLMs to generalize across such diverse downstream tasks in context are not yet fully understood, posing a challenge for the reliability and scientific use of LLMs. Physical systems inherently provide structured, interpretable, and easily controllable data, making physics-based tasks an ideal domain for addressing this challenge. Such well-defined problems serve as a bridge between synthetic, abstract benchmarks and more realistic settings, advancing our understanding of LLMs toward broader applicability in the physical world.

Recent studies have begun to uncover the mechanisms behind ICL by identifying specialized attention head structures in LLMs~\cite{hendel2023incontext, merullo2024language, todd2024function, olsson2022context}.
A complementary line of work extracts interpretable features from LLM neurons using sparse autoencoders (SAEs)~\cite{bricken2023monosemanticity, templeton2024scaling, huben2024sparse}, and recent efforts have applied SAEs to study disentangled features specifically in the context of ICL~\cite{demircan2025sparse, kharlapenko2025scaling}.

\begin{figure*}[!t]
    \centering
    \includegraphics[width=\textwidth]{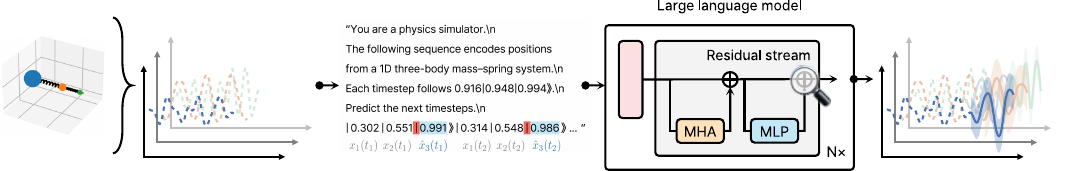}
    \caption{An overview of our procedure. From the simulated dynamical system, we prepare trajectory data and an input prompt that asks the model to forecast the future of the given system. During the inference phase, we track the residual stream activations at the specific token positions (highlighted in red in the prompt) that precede the generated output tokens (highlighted in blue). We then analyze the collected residual stream activations.
    }\label{fig1}
\end{figure*}

In this work, we bridge ICL and mechanistic interpretability to investigate whether LLMs implicitly learn physics in context. We prompt a pre-trained LLM to forecast the future dynamics of physical systems, and analyze its internal activations to assess whether physics-related features are encoded inside the model.

The central question is: \emph{Do recognizable physical quantities (e.g., displacement, velocity, or energy) emerge as distinct features within an LLM's internal representations during in-context learning?}
Answering this question would not only illuminate LLMs' reasoning strategies but also test whether language models spontaneously discover human-understandable physics concepts when needed for a task, providing a novel glimpse into how physical structure is organized inside LLMs during ICL.
An overview of our procedure is provided in Figure~\ref{fig1}. \newline

Our contributions are summarized as follows.
\begin{itemize}
    \item We demonstrate that pre-trained LLMs can predict the dynamics of physical systems via in-context learning, with prediction accuracy improving with the context length.
    \item We identify energy-correlated features in the residual streams of LLMs, that captures physically meaningful concepts rather than mere numerical patterns.
    \item We provide evidence that energy latents are robust across context length, system complexity, and model scale.
    \item We introduce a gradient-based attribution analysis and find that the model's prediction concentrates on energy-correlated features during in-context forecasting.
\end{itemize}

\section{Preliminaries}
\label{sec.2}
\subsection{Learning physics of a system with neural networks}
\label{sec.2.1}
This line of study stems from a body of work that parameterizes grounded physical principles or governing functionals such as Hamiltonian or Lagrangian with neural networks~\cite{greydanus2019hamiltonian, cranmer2020lagrangian, wetzel2020discovering, ha2021discovering, song2024towards}.
More recently, transformer-based approaches have rapidly developed for forecasting a broad range of time series, including physical dynamics, either by developing task-specific architectures~\cite{chen2023tsmixer, nie2023a, liu2024itransformer} or by leveraging pre-trained LLMs~\cite{xue2023promptcast, gruver2023large, jin2024timellm}.

We note that physical systems are particularly attractive because they produce interpretable data, and exhibit behaviors from simple oscillations to rich, complex dynamics.
They also provide controlled environments in which the representations learned by models can be effectively probed and interpreted.
This makes them particularly suitable for studying the emergent reasoning behaviors of LLMs in a realistic yet tractable setting.

\subsection{LLM's ability to learn in-context}
\label{sec.2.2}
Learning in context, or ICL, refers to the ability of LLMs to perform tasks based solely on the examples provided in the prompt, without any parameter updates. This capability far exceeds the original objective of LLMs, which was to predict the next token in large text corpora~\cite{brown2020language, wei2022emergent, bubeck2023sparks}.
In particular, LLMs have been shown to predict time series through ICL via tokenization schemes that map sequences of floating-point numbers to text inputs suitable for LLMs~\cite{gruver2023large}.

Understanding how LLMs achieve ICL on these various tasks remains an open question. From a theoretical perspective, several works proposed that compact representations of tasks (e.g.  task vectors, function vectors) exist within a subset of attention heads~\cite{hendel2023incontext, merullo2024language, todd2024function}. Other works identified \emph{induction heads}, specialized attention structures that complete patterns by attending to previously occurring sequences~\cite{elhage2021mathematical, olsson2022context}. However, predicting physical dynamics likely requires more than mere pattern matching; the model may also need to internally infer the laws of motion governing the data. A recent study by~\citet{liu2024llms} has hinted that LLMs do form such implicit models, demonstrating that LLMs can accurately recover the transition rules underlying various dynamics whose evolution is driven by physical principles. This suggests that LLMs might generalize physical principles in context. In our experiments, we leverage this in-context capability by providing trajectories of toy physical systems for the model to forecast, and then probe the hidden activations that accompany its predictions.

\subsection{Interpretable features to world models in LLMs}

\label{sec.2.3}
Interpreting the latent representations of LLMs (or, more broadly, artificial neural networks) remains a core challenge for AI research. One significant obstacle is the prevalence of \emph{polysemanticity} (individual neurons often encode multiple, entangled features) which complicates direct interpretability efforts~\cite{olah2020zoom, elhage2022superposition}. Recently, SAEs have emerged as an effective remedy: building on sparse dictionary learning algorithms~\cite{olshausen1997sparse, lee2006efficient}, SAEs reconstruct model activations using a sparse combination of latent features, effectively decomposing complex representations into a dictionary of monosemantic, interpretable features or circuits~\cite{yun2021transformer, sharkey2023technical, bricken2023monosemanticity, templeton2024scaling, huben2024sparse, marks2025sparse}. Furthermore, recent work has begun to apply SAEs to LLMs to elucidate the mechanisms behind ICL in simple settings such as reinforcement learning and task-vector tasks~\cite{demircan2025sparse, kharlapenko2025scaling}.

Another line of work investigates whether transformer-based foundation models trained on task-specific data genuinely learn the underlying \emph{world model}. \citet{vafa2025what} argued that a transformer trained on orbital trajectory data fails to apply Newtonian mechanics, which would constitute the underlying world model, when adapting to new physics tasks.
A follow-up by~\citet{liu2026kepler} suggested that introducing minimal inductive biases can narrow the gap between such foundation models and a faithful world model.
However, these discussions have focused on GPT-2-scale transformers trained on specific task data.
In this work, we instead seek to uncover features which corresponds to a world model in the sense of the prior literature, inside LLMs that enable them to learn physics in context.

\section{Methods}
\label{sec.3}
This section details the experimental procedures for evaluating in-context physics learning in LLMs and for investigating their internal representations.
We first describe the ICL procedure for our physics-based task, including data generation and the inference process, and then describe the methods used to uncover meaningful internal representations.

\subsection{Testing LLMs solve physics tasks in-context}
\label{sec.3.1}
\noindent\textbf{Physics task preparation}\hquad To evaluate whether LLMs can learn physics in context, we task them with predicting the dynamics of physical systems.
By default, we focus on a 1D coupled mass-spring system with three masses, which exhibits structured but non-trivial dynamics.

We simulate the system to generate time-series trajectories of the three masses, $\bm{x}(t_i)=(x_1(t_i), x_2(t_i), x_3(t_i))$, where $x_j(t_i)$ denotes the position of the $j$-th mass at time step $t_i$, sampled at a discrete time interval $\Delta t = 0.1$.
Details of the data generation procedure are described in the Supplementary Material (SM)~\ref{sec.a}.

\noindent\textbf{Tokenizing the trajectories}\hquad Given the simulated trajectories, we pre-process the array of trajectories into tokenized inputs following the procedure of~\citet{gruver2023large}, which converts numerical values into sequences of number tokens. 

There exist several strategies for tokenizing numbers in LLMs, including single-digit, left-to-right or right-to-left 3-digit, and mixed-digit (i.e., byte-pair encoding) tokenization.
Prior studies have shown that single-digit tokenization often yields better performance on arithmetic tasks than alternative schemes~\cite{liu2023goat, singh2024tokenization, lee2024from, yang2025number}.
Consistent with these findings, many recent open-sourced LLMs (e.g. Gemma 3, Qwen3, and Ministral 3) adopt single-digit tokenization schemes. In this work, we take Qwen3 as a representative example of an LLM employing single-digit tokenization, and use the Qwen3-14B model as our default. Details of the tokenization procedure are described in the SM~\ref{sec.b.1}.

\noindent\textbf{Inference on LLMs}\hquad After preparing the tokenized sequences, we prompt the model to generate future positions, prefixed by a natural-language description that specifies the forecasting setting (i.e., the physical system and the input/output format). Following this description, we represent the three-mass state at each time step as a single textual \emph{chunk}, in which masses are separated by a mass delimiter ``\texttt{|}'' and each chunk is terminated by a time delimiter ``\texttt{>>}''.
For example, one time step is encoded as
\texttt{\string|0.531\string|0.123\string|0.912\string>\string>},
corresponding to $x_1(t_i)$, $x_2(t_i)$, $x_3(t_i)$ at time step $t_i$.
We then prompt the model with a sequence of such chunks, one time step at a time, following common practices for LLM-based time-series forecasting~\cite{gruver2023large, nie2023a, liu2024autotimes, jin2024timellm, cao2024tempo}. The exact prompts used in our experiments are provided in the SM~\ref{sec.b.2}.

To isolate a consistent prediction point for representation analysis and simplify the forecasting objective, we employ a \emph{partially autoregressive} decoding scheme. In this setup, $x_1$ and $x_2$ are provided as ground truth at every time step, while $x_3$ is generated by the model.
Specifically, at each future time step we inject a ground-truth prefix containing the delimiter tokens together with the values of $x_1$ and $x_2$, and then generate only the numerical tokens corresponding to $x_3$.
For example, given an input context ending with
\texttt{[Prompt]\ldots\string|0.853\string|0.914\string|0.986\string>\string>},
the first two prediction steps can be written as
\[
\begin{array}{rcc}
& \text{injected prefix} & \text{generated value} \\
t_1: &
\texttt{\string|0.855\string|0.915\string|} &
\texttt{0.941} \\
t_2: &
\texttt{\string|0.858\string|0.917\string|} &
\texttt{0.946}
\end{array}
\]
The time delimiter ``\texttt{>>}'' is appended between consecutive state chunks.
Thus, throughout the rollout, the ground-truth $x_1$ and $x_2$ coordinates are injected at every prediction step, while only the $x_3$ coordinate is generated autoregressively. We repeat this procedure to obtain a multi-step rollout for $x_3$.

We evaluate forecasting performance across various context lengths $L_{\mathrm{hist}} \in \{32, 64, 128, 256, 512\}$, where $L_{\mathrm{hist}}$ is the number of history steps included in the prompt, and task the model with forecasting the subsequent 32 time steps.
The hyperparameters used for inference are provided in the SM~\ref{sec.b.3}.

\begin{figure*}[t!]
    \centering
    \includegraphics[width=\textwidth]{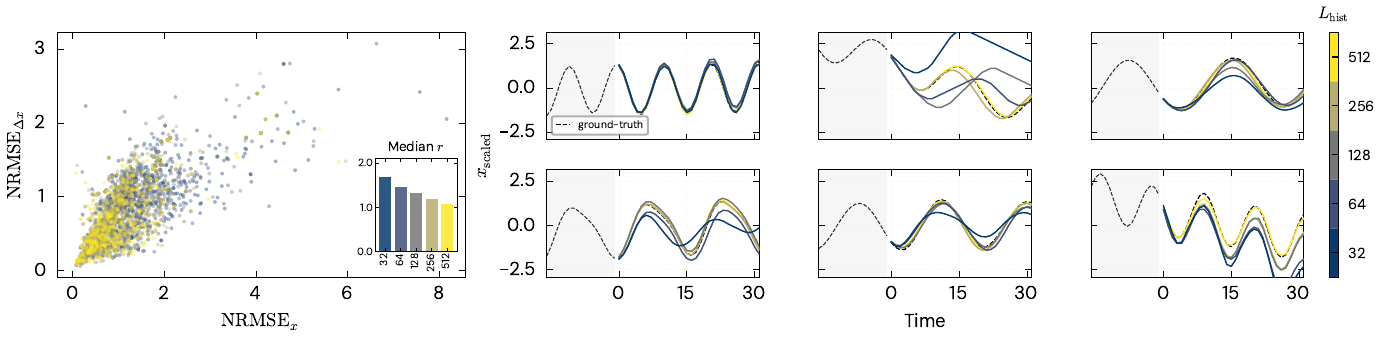}
    \caption{Prediction performance of Qwen3-14B under different in-context history lengths. 
(Left) Forecasting errors are shown in the two-dimensional error plane defined by the displacement NRMSE and the velocity (finite-difference) NRMSE. Each point corresponds to one rollout, and colors indicate the history length $L_{\mathrm{hist}}$. The inset shows the median composite forecast error.
(Right) Representative prediction examples for different $L_{\mathrm{hist}}$. The displacement trajectories are standardized for visualization. The shaded region indicates the final portion of the observed history shown before the forecast window.
    }\label{fig2}
\end{figure*}

\subsection{Investigation of residual stream in LLMs}\label{sec.3.2}
\noindent\textbf{Residual stream analysis}\hquad Interpretability analyses commonly target the transformer residual stream because attention and MLP submodules communicate through this shared representation space~\cite{elhage2023privileged, gurnee2023finding}. Building on this view, prior works have sought to extract interpretable concepts either directly from residual activations or from SAE latents trained on them~\cite{sofroniew2026twheemotion, huben2024sparse, templeton2024scaling, demircan2025sparse}.

In this study, we primarily analyze the raw residual stream activations rather than SAE features. This choice is motivated under two considerations. 
(i) Even with a well-trained SAE, faithfully reconstructing the residual stream is essential, as the model prediction should ideally remain unchanged after reconstruction. This requirement is particularly stringent here because the float-valued output of interest can change under even small perturbations. Since achieving effectively zero reconstruction error with an SAE is difficult, we primarily analyze the raw residual stream. (ii) Moreover, physics-relevant information may already be represented in a relatively disentangled, less polysemantic form. For completeness, we also report the corresponding SAE-based analysis in the SM~\ref{sec.c.1}.

\noindent\textbf{Extraction point in the prompt}\hquad Our analysis targets the internal state immediately preceding the generation of $x_3$.
For each prediction step, we extract the residual stream vector at the token position corresponding to the mass delimiter ``\texttt{|}'' that precedes the $x_3$ field to be generated.
Collecting this position across the rollout yields a sequence of residual vectors of shape $[\tilde{L}, H]$ at each transformer block, where $\tilde{L}$ is the number of prediction steps (i.e., the number of extracted delimiter positions) and $H$ is the model hidden dimension.

\noindent\textbf{Searching for meaningful concepts}\hquad Conventional residual stream analyses leverage dictionary-vector interpretation for semantic concept discovery. However, this approach is challenging when applied to raw physical-system data that lacks linguistic anchors. We therefore focus on whether physically meaningful quantities manifest through statistical associations between internal activations and physical scalars.
Specifically, if the model internally tracks physical quantities such as energy, this should be reflected in correlations between (i) raw residual-stream components and physical quantities, and/or (ii) sparse activations produced by SAEs and physical quantities.

\noindent\textbf{Correlation analysis}\hquad At block layer $l$ we compute correlations between internal activations extracted at the mass delimiter token ``\texttt{|}'' that precedes $x_3$ and physical quantities of interest.
Let $\vec{h}^l(t_i)\in\mathbb{R}^{H}$ denote the extracted residual-stream vector at time step $t_i$ (or $\vec{h}^l(t_i) \in \mathbb{R}^{aH}$ for sparse activations, where $a$ is the SAE expansion factor).
We pair each activation with a physical scalar $\mathcal{Q}(t)$ (e.g., total energy $E(t)$, kinetic energy $T(t)$, or potential energy $V(t)$) and aggregate pairs across trajectories and time steps.

For each activation $h_j^l$, we compute the absolute Pearson correlation
\begin{equation}
\rho_p(h_j^l, \mathcal{Q}) \;=\; \Biggl| \cfrac{\mathrm{cov}(\{h_j^l\}, \{\mathcal{Q}\})}{\sigma_{\{h_j^l\}}\sigma_{\{\mathcal{Q}\}}}\Biggr|.
\end{equation}
A high correlation $\rho_{p}(h_j^l,\mathcal{Q})$ indicates that the corresponding latent feature is associated with the physical quantity $\mathcal{Q}$, whereas a low value indicates little such association.

\section{Results}\label{sec.4}
\subsection{Effects of history length during ICL}\label{sec.4.1}

We first evaluate how the amount of in-context history affects the forecasting performance of Qwen3-14B on the 3-body mass-spring system. Since the target trajectories are periodic and oscillatory, a point-wise error on the displacement alone can be misleading. For example, a degenerate prediction that collapses toward the mean trajectory may achieve a moderate displacement error while failing to reproduce the oscillatory dynamics. We therefore evaluate predictions using two complementary normalized errors: one for the displacement and one for its velocity, computed via finite differences.

\begin{figure*}[t]
    \centering
    \includegraphics[width=\textwidth]{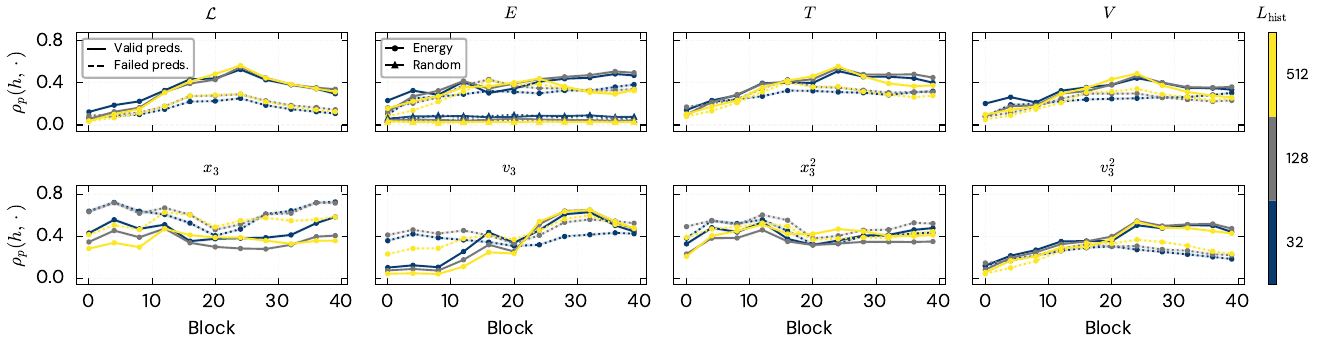}
    \caption{Block-wise top-30 absolute correlations between residual stream activations and physics-related quantities (Lagrangian $\mathcal{L}$, total energy $E$, kinetic energy $T$, potential energy $V$, displacement $x_3$, $x_3^2$, and velocity $v_3$, $v_3^2$) across different history length $L_\text{hist}$.
    In the $\rho_p(h, E)$ plot, circle markers denote $\rho_p(h, E)$ and square markers denote $\rho_p(h, R)$, where $R$ is a random arbitrary value drawn at the same scale as $E$. Solid and dotted lines correspond to correlation values computed from the residual streams of valid and failed inferences, respectively.
    }\label{fig3}
\end{figure*}

For a predicted trajectory $\hat{x}$ and the corresponding ground-truth trajectory $x$, we define the displacement error as
\begin{equation}
\mathrm{NRMSE}_{x}
=
\frac{
\sqrt{\frac{1}{T}\sum_{t=1}^{T}(\hat{x}_{t}-x_{t})^{2}}
}{
\sigma(x) + \epsilon
},
\end{equation}
where $\sigma(x)$ denotes the standard deviation of the ground-truth displacement and $\epsilon$ is a small constant for numerical stability. To further penalize predictions that match the average displacement but fail to capture the local dynamics, we also compute a velocity error using the first temporal difference, 
\begin{equation}
\mathrm{NRMSE}_{\Delta x}
=
\frac{
\sqrt{\frac{1}{T-1}\sum_{t=1}^{T-1}
\left(
\Delta \hat{x}_{t}-\Delta {x}_{t}
\right)^{2}}
}{
\sigma(\Delta x) + \epsilon
},
\end{equation}
where $\Delta x_t = x_{t+1}-x_t$. This second term acts as a finite-difference proxy for velocity error and captures whether the model reproduces the phase and local variation.

To summarize the two errors with a single scalar, we define the composite normalized error as an overall metric to evaluate our forecasting task:
\begin{equation}
r =
\sqrt{
\mathrm{NRMSE}_{x}^{2}
+
\mathrm{NRMSE}_{\Delta x}^{2}
}.
\end{equation}
This metric corresponds to the Euclidean distance from the origin in the two-dimensional error plane, where smaller values indicate better agreement with the ground-truth trajectory in both displacement and temporal variation.

Figure~\ref{fig2} shows the forecasting results across history lengths $L_{\mathrm{hist}} \in \{32,64,128,256,512\}$. In the error plane, predictions with longer context lengths concentrate increasingly toward the lower-left region, indicating that both $\mathrm{NRMSE}_{x}$ and $\mathrm{NRMSE}_{\Delta x}$ decrease as more past trajectory information is provided. The qualitative examples show the same trend: short-context predictions often capture only coarse oscillatory patterns, whereas longer-context predictions more accurately follow the phase and amplitude of the ground-truth dynamics.
Together, these results demonstrate that Qwen3's performance on the in-context forecasting of physical systems improves with the in-context history length, both qualitatively and quantitatively.

\subsection{Correlations of latents with physical quantities}\label{sec.4.2}

To further investigate the ICL capability demonstrated in Section~\ref{sec.4.1}, we analyze the residual streams of Qwen3 during inference.

We hypothesize that, in the context of physics-based tasks, the model represents meaningful physical concepts in its internal activations, beyond simple pattern matching. A natural candidate is energy, which plays a central role in dynamical systems governed by Hamiltonian or Lagrangian mechanics, where energy conservation emerges as a fundamental principle~\cite{goldstein2001classical}. 

Following the methodology described in Section~\ref{sec.3.2}, we track the residual streams across the transformer blocks of the Qwen3-14B model (specifically at block layer $l \in [0, 4, 8, \ldots, 36, 39]$). For each block, we collect features from inference runs using different context lengths.

\begin{figure}[h!]
    \centering
    \includegraphics[width=\linewidth]{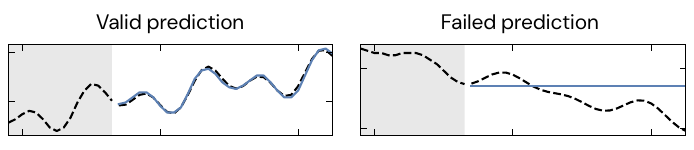}
    \caption{Examples of model predictions. The dotted line, the gray area, and the blue line each represents the ground-truth trajectory, the history given, and the predictions respectively.
    }\label{fig4}
\end{figure}

Here, we compare the residual streams in two regimes: valid inferences and failed inferences. Examples are shown in Figure~\ref{fig4}.

Even though the same prompt is fed to the model, there exist extreme cases in which the prediction fails drastically for some trajectory samples; these will serve as a control group throughout our analyses.

Figure~\ref{fig3} presents a block-wise view of the highest correlation values between residual stream activations and total energy across all transformer blocks.
By looking into the block-wise distribution of $\rho_p(h, \mathcal{Q})$ (we drop $j,l$ indices for convenience) in Figure~\ref{fig3}, we can qualitatively assess where in the model energy-related features are most strongly represented. The figure shows, for each block, the highest correlation observed between any activation coordinate and a given physical quantity.

\begin{figure*}[t!]
    \centering
    \includegraphics[width=\textwidth]{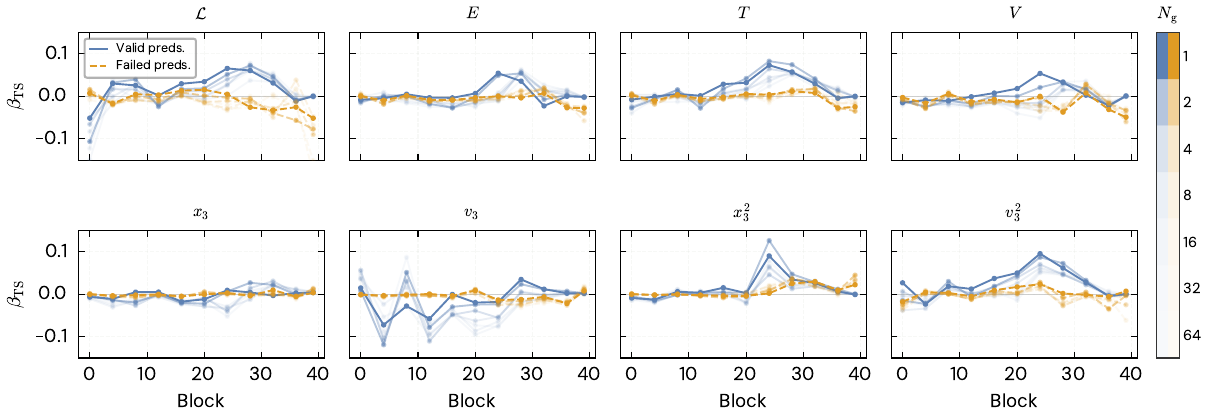}
    \caption{Block-wise alignment between correlation score and gradient attribution of physical quantities (total energy $E$, Lagrangian $\mathcal{L}$, kinetic energy $T$, potential energy $V$, displacement $x_3$, $x_3^2$, and velocity $v_3$, $v_3^2$) across different group size $N_\text{g}$. Solid and dotted lines correspond to values computed from the residual streams of valid and failed inferences, respectively.
    }\label{fig5}
\end{figure*}

We first examine the correlation with the Lagrangian of the system, $\rho_p(h, \mathcal{L})$ (top-left subplot of Figure~\ref{fig3}). The early blocks exhibit relatively weak correlations, but the correlations grow as we move toward the middle blocks. Beyond the middle blocks, the correlation plateaus or even slightly decreases, suggesting that energy-related information is most saliently captured in the model's mid-level representations.

Next we dissect the correlation with total energy (i.e., the Hamiltonian in our case). We compare it against a random-value baseline $\rho_p(h, R)$, where the random values $R$ are sampled at the same scale as the total energy (top row, second column of Figure~\ref{fig3}). Energy yields substantial correlations with residual stream activations across all transformer blocks, while the random baseline does not.
% \todo{Since total energy is a conserved quantity, thus it is constant throughout token positions,
We substantiate that the existence of this correlation to energy is not due to random chance by capturing numbers with similar scale. This contrast indicates that residual stream activations carry information aligned with energy, suggesting that the model already maintains internal representations of physical concepts.

To assess whether other energy-related quantities are similarly represented, we also compute the correlation between model activations and kinetic energy $T$, potential energy $V$ (top row, third and fourth columns of Figure~\ref{fig3}). These results provide further evidence that the residual streams in specific blocks correlate significantly with these quantities. 
For all these quantities ($T$, $V$), we observe a similar trend observed from Lagrangian $\mathcal{L}$. This pattern implies that the middle layers play a crucial role in encoding physical quantities, after which the model may shift to integrating this information with other features relevant for downstream prediction.

The emergence of energy-related representations is not abrupt at longer history lengths; instead, the strength of these correlations gradually intensifies as the context length increases. This suggests that while some energy-related features can be captured even at shorter contexts, their prominence and reliability are enhanced as the model processes longer histories.

It is also worth noting that $\rho_p(h, \cdot)$ for valid predictions is consistently higher than for failed predictions in the cases of $T$, $V$, and $\mathcal{L}$. One might expect $\rho_p(h, E)$ to behave similarly, but this is less pronounced because we use conservative dynamics, in which the total energy is constant in time and therefore yields naturally lower Pearson correlation than time-varying signals. This observation suggests that physical concepts are more cleanly formed inside the model when its dynamics predictions are more accurate.

Beyond the direct energy-related quantities, we also investigate intermediate quantities such as $x_3$, $v_3$, $x_3^2$, and $v_3^2$ (lower row of Figure~\ref{fig3}). Unlike the direct energy quantities, $\rho_p(h, x_3)$ and $\rho_p(h, x_3^2)$ are higher for failed predictions than for valid ones, and these correlations form an overall plateau across the transformer blocks rather than peaking at the mid blocks. In addition, $\rho_p(h, x_3)$, $\rho_p(h, v_3)$, and $\rho_p(h, x_3^2)$ for the failed cases remain consistently high throughout all transformer blocks. This suggests that, in the failed-inference regime, the model encodes surface-level patterns more strongly than the underlying physical concepts.
We note, however, that the failed predictions we identify (Figure~\ref{fig4}) typically correspond to a degenerate collapse toward a near-constant output. We therefore cannot fully rule out that the elevated $\rho_p(h, x_3)$ and $\rho_p(h, x_3^2)$ in this regime reflect this specific failure mode rather than a generic property of unsuccessful in-context inference. Characterizing other failure modes---e.g., phase-shifted or amplitude-distorted predictions---is left for future work.

We also extract sparsified features using SAEs and apply the same correlation analysis. The results show no notable advantage for the reasons discussed previously; we therefore retain the analysis based on raw residual stream activations. We report the corresponding SAE-based results in the SM~\ref{sec.c}.

\subsection{Varying system complexity and model scale}\label{sec.4.3}
\noindent\textbf{System complexity}\hquad
We extend our analysis to systems of differing complexity: a 2-body mass-spring system, which exhibits simple oscillatory behavior, and a 4-body mass-spring system, which exhibits highly complex dynamics.
As shown in Figure~\ref{figS3}, both the 2-body and 4-body systems exhibit weak correlations at the early blocks, peak correlations at the middle blocks, and decreasing correlations toward the deeper blocks, consistent with the previous results.

\noindent\textbf{Model scale}\hquad
Finally, we conduct the correlation analysis across a wide range of model scales, from 0.6B to 14B parameters. Figure~\ref{figS4} shows block-wise $\rho_p(h, \mathcal{L})$ for different models. The trend of peak correlation at mid-block depth persists consistently across model sizes. Notably, the maximum $\rho_p(h, \mathcal{L})$ at the mid blocks increases as the model scales up. This observation suggests that the model's capacity to maintain physical concepts as internal representations strengthens with scale.

\subsection{Gradient-based attribution analysis}
\label{sec.4.4}
We now ask whether residual stream activations with high $\rho_p(h, \mathcal{Q})$ play a functional role during in-context dynamics forecasting. One option would be to craft steering vectors from the most strongly correlated residual dimensions. Physical concepts, however, are neither binary nor categorical, unlike those typically studied in natural language, and it remains unclear how to construct latent directions that steer the model along a target physical axis~\cite{joseph2026interpreting}. We therefore propose a gradient-based attribution analysis that measures, at an aggregate level, how well the identified features align with the forward-pass computation the model actually performs.
 
At block layer $l$, we rank the residual stream activations in descending order of $\lvert\rho_p(h_j^l, \mathcal{Q})\rvert$ for the target physical quantity $\mathcal{Q}$, and coarse-grain them into subgroups of size $N_\text{g}$.
We then represent the correlation score $C_J^l$ of group $J$ as the mean of $\lvert\rho_p(h_j^l, \mathcal{Q})\rvert$.
For each activation dimension within a group, we compute the gradient-activation product $G_j^l\odot h_j^l$, where $G_j^l$ is the gradient of the summed $x_3$ digit log-probability with respect to the residual stream activation $h_j^l$. We can now define the gradient attribution score $A_J^l$ of each group $J$ by taking the median value of $G_j^l\odot h_j^l$.
To quantify how gradient attribution is aligned to latents correlated with $\mathcal{Q}$, we compute the Theil-Sen slope $\beta_\textrm{TS}$ of the linear fit between $C_J^l$ and $A_J^l$.
We provide further details of this procedure in the SM~\ref{sec.d}.
 
Figure~\ref{fig5} reports $\beta_\textrm{TS}$, a single scalar relating (i) how strongly a group of activations linearly encodes a chosen physical quantity and (ii) how much that same group contributes to the model's numerical prediction in context.
Here, a positive $\beta_\textrm{TS}$ indicates that residual groups more strongly associated with the physical quantity also tend to have larger absolute gradient attribution to the numerical prediction log-probability.
A negative $\beta_\textrm{TS}$ indicates the opposite, with attribution concentrated in groups whose physical correlations are weaker.
A value near zero indicates no clear linear relationship between the two group profiles.

We observe that, for valid predictions, the largest positive slopes appear in the intermediate layers, where the strongest physical correlations were also observed. For the case of failed predictions, the slopes generally remain near zero.
Additionally, the valid-prediction result for $v_3$ shows large fluctuations across the lower blocks. As shown in Figure~\ref{fig3}, the lack of strongly correlated residual features makes the correlation-based grouping effectively random.
Residual groups that strongly represent a physical quantity do not receive systematically larger attribution when the prediction fails. 
This suggests that, for valid predictions, activations strongly correlated with a physical quantity also contribute functionally to the numerical prediction.

By contrast, input-adjacent variables such as $x_3$ and $v_3$ show stronger residual correlations for failed predictions than for valid ones in Figure~\ref{fig3}, yet generally yield near-zero slopes for both valid and failed predictions in Figure~\ref{fig5}. We therefore interpret their lower-layer representations as passive reflections of the input rather than features actively used for prediction.
In turn, higher-level quantities like energy, appear to be represented more selectively and to contribute more strongly to successful predictions.
These results indicate that the emergent physical representations we identified are not merely decodable correlates of the input, but are functionally relevant to the model's in-context forecasting computation, with the degree of participation tracking both the abstraction level of the quantity and the depth at which it is encoded.

\section{Discussion}
The ability of pretrained language models to predict the trajectories of physical systems highlights their remarkable in-context learning capabilities. Our analysis reveals that prediction accuracy improves with longer context windows, and that this improvement coincides with the emergence of internal representations correlating with physical quantities. This suggests that LLMs may spontaneously encode governing physical concepts rather than relying solely on surface-level pattern extrapolation.

We summarize the findings supporting this conclusion. First, pretrained LLMs exhibit in-context learning of nonlinear physical dynamics, with prediction accuracy improving as context length increases. Second, the emergence of energy-correlated representations is context-dependent: residual stream activations show substantial correlations with physics-related quantities.
Furthermore, these correlations are robust across context lengths, system complexities, and model scales.

A key limitation of our study is the use of raw trajectory values as the sole input modality, without external instructions or prompting strategies. It remains an open question whether incorporating scratchpads, chain-of-thought reasoning, or structured prompts~\cite{xue2023promptcast, lee2024teaching} would further strengthen or clarify the observed correlations. Additionally, while we adopt trajectory forecasting as a controlled probe of physical reasoning, alternative task formulations may offer complementary insights.
Future work could explore physical concepts beyond classical mechanics, or investigate more human-aligned tasks, such as answering conceptual questions drawn from physics exams.
Other limitation is that our valid/failed comparison relies on a particular failure mode (collapse to a near-constant output), as discussed in Section~\ref{sec.4.2}, the generality of these correlations beyond this failure mode remains to be further investigated.

In summary, our results extend the frontier of in-context learning from symbolic and linguistic tasks to structured physical dynamics. The model's capacity to form representations aligned with energy, which is a foundational inductive bias in physics, from raw time-series inputs suggests that transformer architectures can internalize abstract reasoning priors. This mirrors human-like abstraction, in which complex dynamics are distilled into conserved quantities.

We hope our findings motivate further directions for future research: (1) leveraging physics-informed tasks to better understand the internal mechanisms of large-scale models, (2) advancing the design of LLM-based agents capable of reasoning and acting in grounded, physically realistic environments, and (3) building toward LLMs that internalize world models of physical systems, as a step toward more reliable automated scientific discovery.

% Do not include acknowledgments in the anonymous version submitted for review.
% For the camera-ready version, uncomment the unnumbered section below; it must
% appear right before the references.
\section*{Acknowledgments}
This study was supported by the Basic Science Research Program through the National Research Foundation of Korea (RS-2025-00514776).

\sloppy
\bibliographystyle{apsrev4-2}
\bibliography{main}

\pagebreak
\onecolumngrid
\newpage
% \appendix
\begin{center}
\textbf{\large Supplemental Material: Uncovering Spontaneous Physics Representations in In-Context Learning}
\end{center}

\setcounter{section}{0}
\setcounter{equation}{0}
\setcounter{figure}{0}
\setcounter{table}{0}

\renewcommand{\theequation}{S\arabic{equation}}
\renewcommand{\thefigure}{S\arabic{figure}}

\section{Details on physical systems used in the experiments}\label{sec.a}
\subsection{System description}\label{sec.a.1}
Below, we provide a description of the 3-body mass-spring system used in our investigation.
Three masses ($m_1, m_2, m_3$) are connected by two springs (with spring constants $k_1, k_2$ and natural lengths $l_1, l_2$) in 1D. The system is governed by the Hamiltonian
\begin{equation*}
    \mathcal{H} = \displaystyle\sum_{i=1}^3 {\mathbf{p}_i^\top \mathbf{p}_i}/2m_i + \displaystyle\sum_{i=1}^2{k_i(l_i - ||\mathbf{x}_i - \mathbf{x}_{i+1}||)^2}/2,
\end{equation*}
where $\mathbf{p}_i = m_i\mathbf{v}_i$.

The 2-body and 4-body mass-spring systems are defined in an analogous manner.
 
% \textbf{Coupled pendulum}\hquad Three masses ($m_1, m_2, m_3$) are suspended parallelly with springs ($k_1, k_2$) connected between the neighbor masses in 3D, where the system is governed by the following Hamiltonian.
 
% \begin{equation*}
%     \mathcal{H} = \displaystyle\sum_{i=1}^3 ({\mathbf{p}_i^\top \mathbf{p}_i/2m_i + gm_ix_{i,2}}) + \displaystyle\sum_{i=1}^2{k_i(l_i - ||\mathbf{x}_i - \mathbf{x}_{i+1}||)^2/2}
% \end{equation*}
 
% Here, $\mathbf{p}_i = m_i\mathbf{v}_i$, and $x_{i, 2}$ is the z-axis element of the position of mass $i$.

\subsection{Dataset generation}\label{sec.a.2}
 
With the systems defined above, we generate synthetic trajectory data by sampling both the physical parameters and the initial states. 
For each system, the masses \(m_i\), spring constants \(k_i\), and rest lengths \(l_i\) are sampled from uniform distributions over pre-specified ranges. 
Given the sampled rest lengths, we construct the corresponding natural configuration of the masses and sample the initial condition 
\(\mathbf{z}(t_0)=(\mathbf{x}(t_0),\mathbf{v}(t_0))\) by perturbing this configuration. 
Thus, each trajectory instance is determined by a particular choice of physical parameters and initial condition.
 
We simulate the resulting Hamiltonian dynamics using an adaptive ODE integrator, following the simulation procedure of~\cite{finzi2020simplifying}. 
The trajectories are sampled at a fixed time interval \(\Delta t=0.1\). While velocities are in principle obtainable from continuous-time derivatives, our forecasting dataset consists of position trajectories only. 
For each trajectory instance, we extract a sequence of total length \(L_{\mathrm{hist}}+32\),
\[
    \mathbf{z}(t_1), \mathbf{z}(t_2), \ldots, \mathbf{z}(t_{L_{\mathrm{hist}}}),
    \mathbf{z}(\tilde{t}_1), \mathbf{z}(\tilde{t}_2), \ldots, \mathbf{z}(\tilde{t}_{32}),
\]
where the first \(L_{\mathrm{hist}}\) states are used as the observed history and the following 32 states are used as the forecasting target.
 
For evaluation, we select a fixed set of 500 independently sampled trajectory instances, each with independently sampled physical parameters and initial conditions, to obtain at least 20 failed predictions.
The resulting 20 trajectories per valid/failed settings are used to evaluate forecasting performance and collecting residual-stream activations.
 
% \begin{figure}[!b]
%     \centering
%     \includegraphics[width=\linewidth]{figures/supp1.pdf}
%     \caption{Epoch-wise evolution of SAE training losses. The left column corresponds to the coupled mass–spring system and the right column to the coupled pendulum. The average total loss and reconstruction loss per epoch are plotted over 10 epochs. Since the sparsity loss remains zero in initial, only losses are plotted only from epoch $3$ onward. With a SAE hidden dimension, $2H$ = $10240$, sparsity loss confirm that the model learns a sufficiently sparse representation.
%     }\label{figS1}
% \end{figure}
 
\section{Details on forecasting inference procedure}\label{sec.b}
\subsection{Tokenizing trajectory data}\label{sec.b.1}
Our tokenization procedure follows the approach of LLMTime~\cite{gruver2023large}. Assuming channel independence across degrees of freedom, we tokenize each univariate trajectory (e.g., $\mathbf{x}_i(t)$ or $\mathbf{v}_i(t)$) independently.
We pre-process each floating-point time series by removing the decimal points and converting the values into string sequences.
Prior to this conversion, we rescale the values such that the $\alpha$-percentile of the rescaled data becomes $1$, with an additional offset $\beta$.
Specifically, given a time series $(z_i(t_0), z_i(t_1), \ldots, z_i(t_T))$, we apply the transformation $z_i(t_j) \rightarrow (z_i(t_j)-b)/a$, where $b=\min_j z_i(t_j)-\beta(\max_j z_i(t_j)-\min_j z_i(t_j))$ and $a$ is the $\alpha$-percentile of $(z_i(t_0)-b, z_i(t_1)-b, \ldots, z_i(t_T)-b)$. We use $\alpha=0.99$ and $\beta=0.3$ as default values.
For all experiments, we use three digits of precision. As an example, if the scaled series is $(-1.348, -0.74, -0.054, 0.582, 1.050, \ldots)$, the resulting string sequence is ``-1348,-740,-54,582,1050,...". We then pass these processed strings to Qwen3's tokenizer, which encodes each digit as a distinct token under the single-digit tokenization scheme.

\begin{figure*}[t!]
    \centering
    \includegraphics[width=\textwidth]{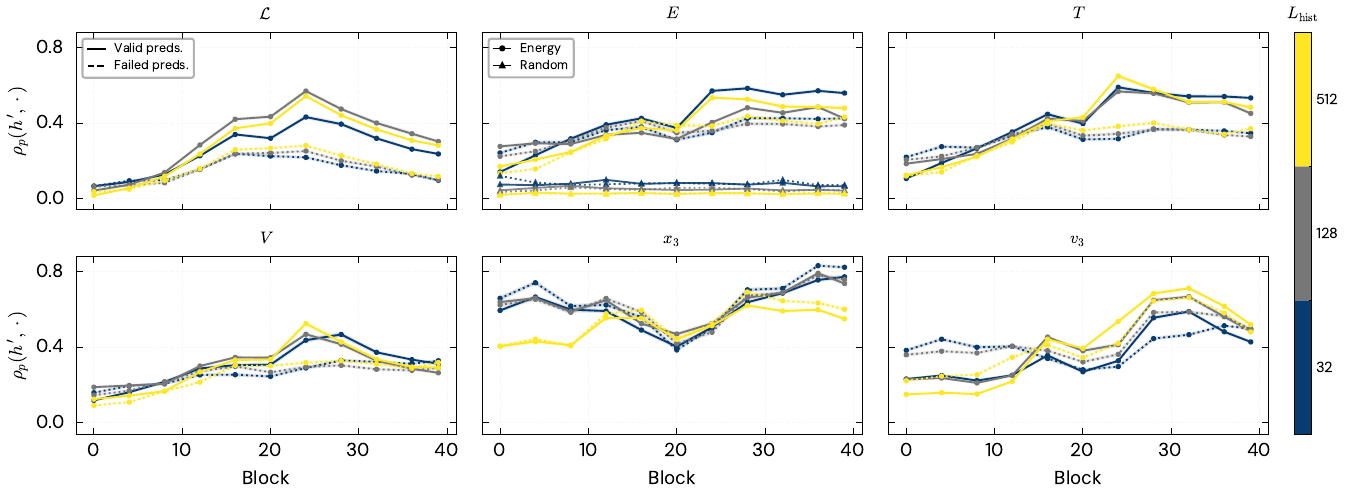}
    \caption{Block-wise top-30 absolute correlations between sparse SAE activations and physics-related quantities across different $L_\text{hist}$: total energy ($E$), Lagrangian ($\mathcal{L}$), kinetic energy ($T$), potential energy ($V$), displacement ($x_3$), and velocity ($v_3$). In the top-left $\rho_p(h, E)$ plot, circle markers denote $\rho_p(h, E)$ and triangle markers denote $\rho_p(h, R)$, where $R$ is a random arbitrary value drawn at the same scale as $E$. Solid and dotted lines correspond to correlation values computed from the residual streams of valid and failed inferences, respectively.
    }\label{figS1}
\end{figure*}
 
\subsection{Prompt examples}\label{sec.b.2}
 
We provide an example prompt used for the 3-body mass-spring system with history length $L_{\mathrm{hist}}=32$ below.
The prompt consists of a short role description, the input-format specification, the forecasting instruction, and the serialized trajectory context.
Line breaks in the trajectory shown below are inserted only for readability; in the actual inference, the full trajectory context is provided as a single serialized sequence of 32 state chunks.

\begin{figure*}[!h]
\centering
\begin{promptbox}[Prompt example for $L_{\mathrm{hist}}=32$]
You are a physics simulator. The following data is encoded time series for position coordinates of mass particles within a 3-body coupled mass-spring system in 1D.
The encoded position coordinate is given in the order x1, x2, x3, and is formatted as in the following example:|0.896|0.925|0.987>>.
Your job is to predict the states for the next time steps based on the given data.
|0.896|0.925|0.987>>|0.897|0.928|0.985>>|0.897|0.932|0.983>>|0.895|0.936|0.981>>
|0.894|0.941|0.978>>|0.893|0.944|0.977>>|0.895|0.946|0.976>>|0.898|0.945|0.976>>
|0.903|0.944|0.978>>|0.907|0.941|0.980>>|0.911|0.938|0.983>>|0.911|0.937|0.986>>
|0.909|0.937|0.989>>|0.904|0.938|0.991>>|0.897|0.940|0.993>>|0.891|0.943|0.995>>
|0.888|0.944|0.997>>|0.887|0.945|0.998>>|0.890|0.945|0.999>>|0.895|0.943|1.000>>
|0.902|0.942|1.000>>|0.908|0.941|0.999>>|0.913|0.942|0.998>>|0.916|0.944|0.996>>
|0.916|0.948|0.994>>|0.914|0.952|0.992>>|0.911|0.956|0.990>>|0.910|0.959|0.989>>
|0.910|0.961|0.990>>|0.911|0.960|0.991>>|0.915|0.957|0.994>>|0.918|0.954|0.997>>
\end{promptbox}
\end{figure*}

\subsection{Inference hyperparameters}\label{sec.b.3}
For inference with Qwen3, we set $\text{temperature}=1$, $\text{top\_k}=50$, and $\text{top\_p}=0.9$.

\section{Additional analysis}\label{sec.c}
\subsection{SAE analysis}\label{sec.c.1}
The SAE employed in our experiments is a two-layer feedforward network with an expansion factor of 2.
Given an input $x \in \mathbb{R}^H$, the encoder computes $c=\mathrm{ReLU}(W_e x + b_e )$, where $W_e \in \mathbb{R}^{2H \times H}$ and $b_e \in \mathbb{R}^{2H}$ are the encoder weight and bias, and $c \in \mathbb{R}^{2H}$ denotes the resulting sparse activations.
A linear decoder $W_d c + b_d$ then produces the reconstruction $\hat{x}$, where $W_d \in \mathbb{R}^{H \times 2H}$ and $b_d \in \mathbb{R}^{H}$ are the decoder weight and bias. Training minimizes the sum of two terms: a mean squared reconstruction error between $\hat{x}$ and $x$, and an $\ell_1$ sparsity penalty on $c$. Concretely, for a mini-batch of size $n=4096$, the total loss is
\begin{equation}
    \mathcal{L}_{\text{total}} = \underbrace{\frac{1}{n}\sum_{i=1}^n\|\hat{{x}}^{(i)}-{x}^{(i)}\|_2^2}_{\mathcal{L}_{\text{recon.}}} \;+ \underbrace{\;\lambda\;\frac{1}{n}\sum_{i=1}^n\|{c}^{(i)}\|_1}_{\mathcal{L}_{s}},
\end{equation}
where $\lambda$ (set to $10^{-3}$) balances reconstruction fidelity against sparsity.
We train 55 distinct SAEs, spanning five history lengths and 11 transformer blocks (block indices $[0, 4, 8, \ldots, 36, 39]$), using the Adam optimizer with a learning rate of $10^{-4}$.
From the trained SAEs, we obtain sparsified activations $\vec{h'}(t_i)\in\mathbb{R}^{2H}$, and then compute the absolute Pearson correlation $\rho_p(h'_j, \mathcal{Q})$ between each sparse coordinate and the physical quantity of interest.

The results are shown in Figure~\ref{figS1}.
The overall trend closely parallels the results based on raw residual stream activations (Figure~\ref{fig3}), confirming that our findings are not an artifact of analyzing raw residual streams.

\subsection{Correlation analysis across system complexity}

In Section~\ref{sec.4.2}, we found that $\mathcal{L}$ exhibits the strongest correlations among the physical quantities considered. We therefore use $\rho_p(h,\mathcal{L})$ as a representative metric to examine how these correlations vary with system complexity.

\begin{figure*}[t!]
\centering
\includegraphics[width=\textwidth]{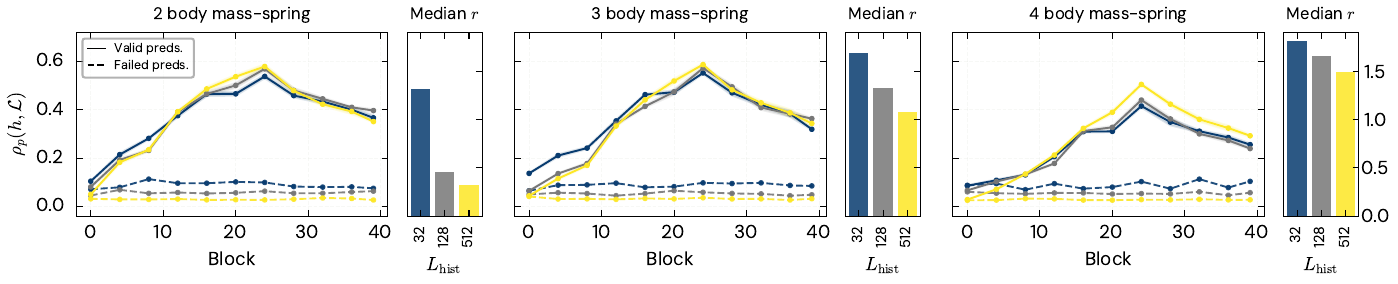}
\caption{
Block-wise top-30 absolute correlations between residual-stream activations and the Lagrangian ($\mathcal{L}$), together with the composite normalized error $r$. Results are shown across context history lengths $L_\text{hist}$ for mass-spring systems with different numbers of bodies.
}
\label{figS2}
\end{figure*}

Figure~\ref{figS2} compares the 2-body, 3-body, and 4-body mass-spring systems. Forecasting performance deteriorates as the number of bodies increases. In contrast, the block-wise correlation profiles do not show a clear monotonic dependence on system complexity. All three systems exhibit weak correlations in the early blocks, stronger correlations at intermediate depths, and decreasing correlations in the deeper blocks. This common profile indicates that the layer-wise organization of the correlated representations is largely preserved across the systems considered.

The 4-body system nevertheless exhibits substantially lower values of $\rho_p(h,\mathcal{L})$ than the 2-body and 3-body systems. In Section~\ref{sec.4.2}, we also found lower correlations for failed predictions than for valid predictions. These observations point to a possible inverse relationship between the strength of the $\mathcal{L}$-correlated representation and the composite normalized error $r$. However, the present comparison does not establish a general trade-off between the two quantities. Systematically testing this relationship across a broader range of system complexities and prediction errors would be an interesting direction for future work.

\begin{figure*}[t!]
\centering
\includegraphics[width=\textwidth]{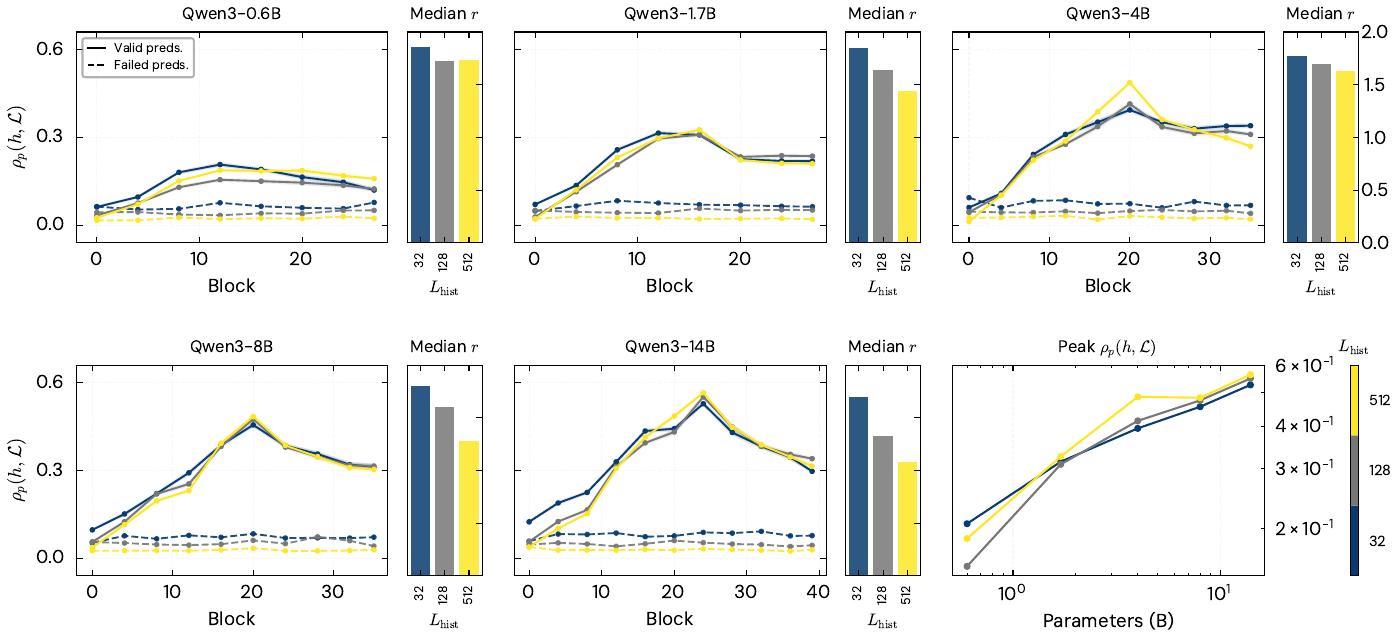}
\caption{
Block-wise top-30 absolute correlations between residual-stream activations and the Lagrangian ($\mathcal{L}$) across Qwen3 models ranging from 0.6B to 14B parameters. The bottom-row, third column summary plot reports the maximum value of $\rho_p(h,\mathcal{L})$ across model depth for valid predictions at each model scale.
}
\label{figS3}
\end{figure*}

\begin{figure*}[t!]
\centering
\includegraphics[width=\textwidth]{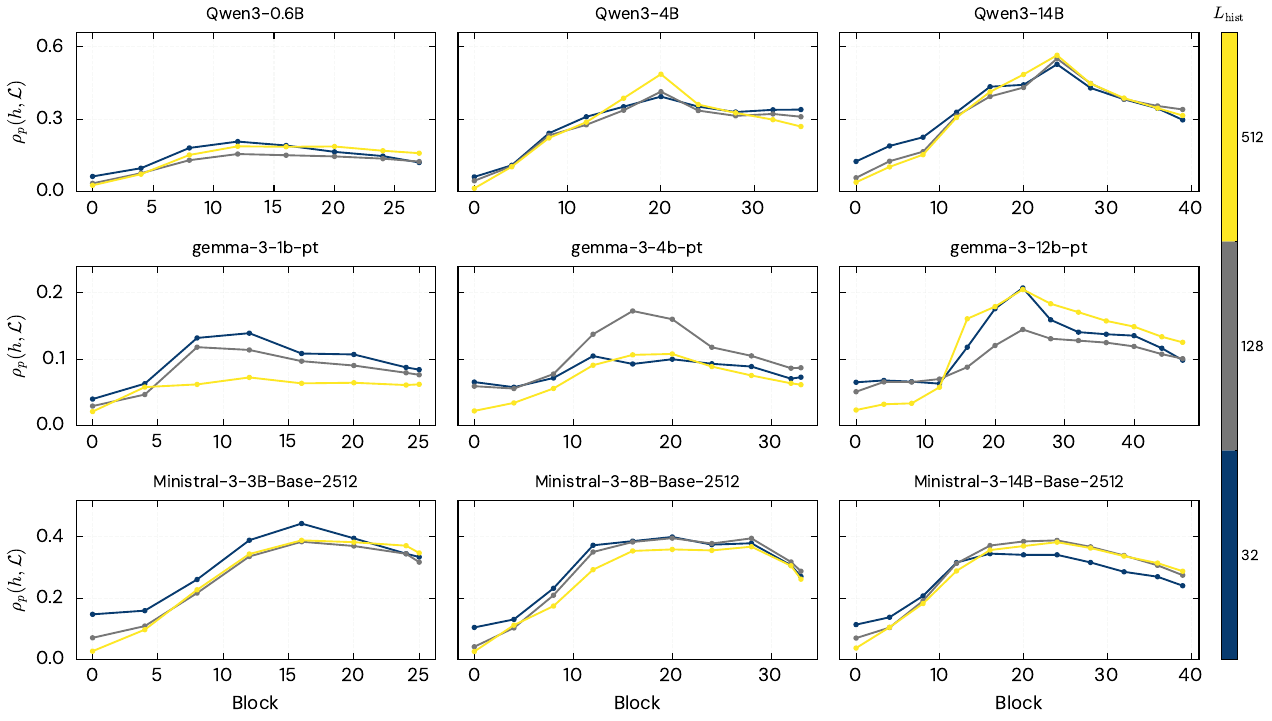}
\caption{
Block-wise top-30 absolute correlations between residual-stream activations and the Lagrangian ($\mathcal{L}$) across different model families and scales.
}
\label{figS4}
\end{figure*}

\subsection{Correlation analysis across model scale and family}

We use $\rho_p(h,\mathcal{L})$ for the 3-body mass-spring system as a representative metric to examine correlations across model scales. As shown in Figure~\ref{figS3}, the block-wise correlation profile peaks at an intermediate depth for every model size tested. As shown in the bottom-row, third-column panel from Figure~\ref{figS3}, the peak value of $\rho_p(h,\mathcal{L})$ increases with model size, suggesting a scaling trend in the maximum correlation strength.

The relative depth of the peak remains approximately constant across model scales. The increase in peak correlation therefore cannot be explained by a shift of the correlated representation to a different relative depth. Instead, the results indicate that the residual-stream representation correlated with $\mathcal{L}$ becomes more pronounced as model capacity increases. This interpretation remains correlational, since a larger value of $\rho_p(h,\mathcal{L})$ does not by itself establish that the representation has a stronger causal role in forecasting.

We further apply our framework to two additional LLM families with single-digit numerical tokenization: Gemma 3 and Ministral 3. As shown in Figure~\ref{figS4}, Qwen3, Gemma 3, and Ministral 3 exhibit qualitatively distinct block-wise correlation profiles, differing in the location, magnitude, and depth dependence of their correlations.

These differences may reflect variations in model architecture, training data, optimization, or other training procedures. For example, Gemma 3 and Ministral 3 use different distillation strategies. However, because the model families also differ in several other respects, the current results cannot isolate the effect of distillation or any other individual design choice. Controlled comparisons that vary one architectural or training factor at a time would therefore be a useful direction for future work to determine what governs the behavior of $\rho_p(h,\cdot)$.

\section{Details on gradient-based attribution analysis}\label{sec.d}

Gradient-based attribution analysis aims to provide a representative quantity whether residual-stream dimensions that are strongly associated with a physical quantity are also important to the model's numerical predictions at each transformer block.
Specifically, the hidden dimensions of each block are ranked by the magnitude of their $\rho_p(h, \cdot)$ value and divided into subgroups.
The following two quantities are then constructed for every group:
\begin{enumerate}
    \item the mean correlation magnitude between residual-stream activations and
    the physical quantity,
    \item the magnitude of the gradient-activation attribution for the numerical
    prediction log-probability.
\end{enumerate}
Each point in the Figure~\ref{fig5} is the slope of the Theil-Sen linear fit of these two group-level vectors. Thus, the ordinate is neither a single-neuron correlation nor a gradient norm. It is a alignment between correlation-profile and gradient-attribution-profile computed from coarse-grained groups within each block.

\noindent\textbf{Experiment setting and notation}\hquad We use the Qwen3-14B model for the gradient-based attribution analysis. Its hidden dimension is $H=5120$, and the prediction length is $\tilde{L}=32$. We consider group sizes $N_\text{g} \in [2^0,2^1,\ldots,2^6]$ and use a history length of $L_\text{hist}=512$ to obtain stable attribution estimates.

% Let $c$ denote a system-setting cohort that corresponds to either one of valid predictions and failed predictions.
% Below, $b=1,\ldots,B$ indexes settings in a cohort, with $B=20$; $t=1,\ldots,P$ indexes autoregressive prediction steps; and $d=1,\ldots,D$ indexes residual-stream coordinates.

\noindent\textbf{Coarse-graining residual stream dimensions}\hquad 
As previously described in Section~\ref{sec.3.2}, we measure the single-neuronal Pearson correlation between the residual stream activation and the target physical quantity of interest $\mathcal{Q}$.
Although we use $v$ as a finite difference that is not divided by $\Delta t$, note that this does not affect the Pearson correlation due to its linearity.

Next, we rank the residual stream dimensions in descending order of $\rho_p(h, \mathcal{Q})$, within each block depth $l$.
Let $\pi^l$ be the permutation that orders the residual dimensions by decreasing $\rho_p(h_j^l,\mathcal{Q})$, such that $\pi^l(r)$ is the dimension index with the $r$-th highest correlation.
% Let $\pi^l(r)$ denote the residual dimension with the $r$-th highest correlation within block (l).
Then, we coarse-grain the ranked residual dimensions to a group-size of $N_\text{g}$.
We denote the $J$-th group activations as, 
\begin{equation*}
    g_J^l
    =
    \left[
    h_{\pi^l\left((J-1)N_{\mathrm{g}}+1\right)}^l,\,
    h_{\pi^l\left((J-1)N_{\mathrm{g}}+2\right)}^l,\,
    \ldots,\,
    h_{\pi^l\left(JN_{\mathrm{g}}\right)}^l
    \right]^{\top},
\end{equation*}
where $J\in\left[1,2,\ldots,\left\lceil H/N_{\mathrm{g}}\right\rceil\right].$
Since $H=5120$ is exactly divisible by any choice of $N_\text{g}$ considered here, no dimensions are discarded.

The correlation score of each group is then defined as
\begin{equation}
C_J^l
=
\frac{1}{N_{\mathrm g}}
\sum_{r\in\mathcal{I}_J}
\left\lvert
\rho_p\left(h_{\pi^l(r)}^l,\mathcal{Q}\right)
\right\rvert,
\label{eq:group-corr}
\end{equation}
where $\mathcal{I}_J = \left\{(J-1)N_{\mathrm g}+1,\ldots,JN_{\mathrm g}\right\}$ denote the set of rank indices assigned to the \(J\)-th group.
Note that this corresponds to the mean absolute correlation, not the mean signed correlation of the group.
Since the groups are built from the $\rho_p(h, \mathcal{Q})$, which takes the absolute value of the Pearson correlation, $C_J^l$ is monotonically-decreasing with group rank $J$ by construction. Group membership is computed
separately for each setting cohort (valid/failed prediction), block and physical quantity.

\noindent\textbf{Gradient-attribution score}\hquad In our partial-autoregressive forecasting, we inject ground-truth $x_1(t)$ and $x_2(t)$ while predicting $x_3(t)$ at every prediction step $t$. 
We denote $\mathcal{D}$ as the set of single digit tokens $q^i(t)$ that constitutes $x_3(t)$. Here, we take the sum of log-probability corresponding to the output $\hat{x}_3(t)$ as,
\begin{equation}
 S(t)=\sum_{q^i(t) \in\mathcal{D}}
 \log p_\theta\!\left(q^i(t)\mid q^{<i}(t),\text{context}\right).
 \label{eq:prediction-score}
\end{equation}
Note that the decimal-point position is excluded from the sum, and the logits are subject to the same disallowed-token mask and temperature adjustment used during generation.

We can now define the gradient-attribution score. Let $\vec{h}^l(t)$ denote the residual stream activation of block $l$ captured by the forward hook at the delimiter ``\texttt{|}'' immediately preceding the $\hat{x}_3(t)$ numerical string.
We first differentiate the prediction score value $S_t$ with respect to $\vec{h}^l(t)$ and then extract the score on every residual dimension, and then take the element-wise product with the residual activation $\vec{h}^l(t)$, resulting in the gradient-attribution score
\begin{align}
 \vec{A}^l(t)
 &=\frac{\partial S(t)}{\partial \vec{h}^l(t)} \odot \vec{h}^l(t) \nonumber \\
 &=\vec{G}^l(t) \odot \vec{h}^l(t).
 \label{eq:gradient}
\end{align}

This quantity is a first-order attribution corresponding to an infinitesimal multiplicative gate on the activation coordinate. It therefore measures the local contribution of each residual-stream dimension to the prediction score. If residual stream dimensions ranked highly by their absolute Pearson correlation with a physical quantity also receive large attribution scores, this suggests that the corresponding representation is functionally relevant and may play a causal role in the model's numerical prediction.

\noindent\textbf{Coarse-grained gradient-attribution}\hquad After computing the gradient-attribution score $A^l$ for all residual dimensions, we revisit the previously defined group $g_J^l$ with indices $\mathcal{I}_J$ and measure the grouped score $A_J^l$ for each J-th group;
\begin{equation}
 A_J^l(t)
 =\mathrm{med}_{r \in \mathcal{I}_J}
 \lvert{A_{\pi^l (r)}^l(t)}\rvert.
 \label{eq:within-group}
\end{equation}

Finally, we aggregate $A_J^l(t)$ by sums it over 20 prediction samples per setting cohort and 32 forecast steps. The resulting grouped gradient-attribution score for every dimension writes as;
\begin{equation}
 A_J^l = \sum_\mathcal{B}\sum_t A_J^l(t),
 \label{eq:group-importance}
\end{equation}
where $\mathcal{B}$ denotes the batch of prediction samples.

\begin{figure*}[t!]
    \centering
    \includegraphics[width=\textwidth]{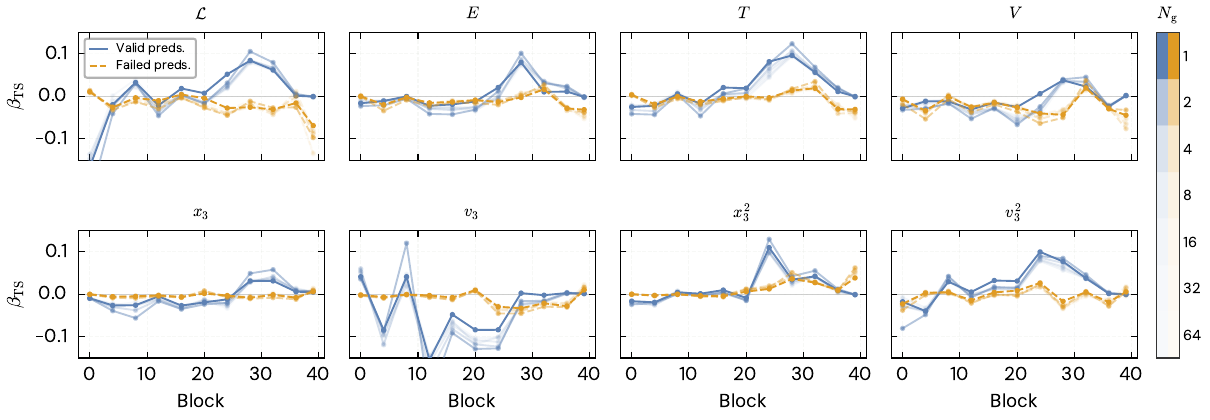}
    \caption{Block-wise alignment between correlation score and gradient attribution (none of the residuals neglected) of physical quantities (total energy $E$, Lagrangian $\mathcal{L}$, kinetic energy $T$, potential energy $V$, displacement $x_3$, $x_3^2$, and velocity $v_3$, $v_3^2$) across different group size $N_\text{g}$. Solid and dotted lines correspond to values computed from the residual streams of valid and failed inferences, respectively.
    }\label{figS5}
\end{figure*}

\noindent\textbf{Correlation-gradient alignment}\hquad For each block $l$, we compute $C_J^l$ and $A_J^l$ for all groups $J$, fit a robust linear relationship between the two measures using the Theil--Sen estimator, and report the resulting slope $\beta_\text{TS}$. Groups with $A_J^l$ above its block-wise 90th percentile are excluded to reduce the influence of extreme values. We additionally report the block-wise alignment results without this exclusion in Figure~\ref{figS5}.

\end{document}